\documentclass{bmvc2k}

\usepackage{multirow}  
\usepackage{arydshln}  
\usepackage{threeparttable}

\title{Learning Not to Reconstruct Anomalies}

\addauthor{Marcella Astrid}{marcella.astrid@ust.ac.kr}{1,2}
\addauthor{Muhammad Zaigham Zaheer}{mzz@ust.ac.kr}{1,2}
\addauthor{Jae-Yeong Lee}{jylee@etri.re.kr}{1,2}
\addauthor{Seung-Ik Lee}{the_silee@etri.re.kr}{1,2}

\addinstitution{
 University of Science and Technology\\
 Daejeon, South Korea
}
\addinstitution{
 Electronics and Telecommunications Research Institute\\
 Daejeon, South Korea
}

\runninghead{Astrid et al.}{Learning Not to Reconstruct Anomalies}

\def\etal{\emph{et al}\bmvaOneDot}

\begin{document}

\maketitle
\begin{abstract}
Video anomaly detection is often seen as one-class classification (OCC) problem due to the limited availability of anomaly examples. Typically, to tackle this problem, an autoencoder (AE) is trained to reconstruct the input with training set consisting only of normal data. At test time, the AE is then expected to well reconstruct the normal data while poorly reconstructing the anomalous data. However, several studies have shown that, even with only normal data training, AEs can often start reconstructing anomalies as well which depletes the anomaly detection performance. To mitigate this problem, we propose a novel methodology to train AEs with the objective of reconstructing only normal data, regardless of the input (i.e., normal or abnormal). Since no real anomalies are available in the OCC settings, the training is assisted by pseudo anomalies that are generated by manipulating normal data to simulate the out-of-normal-data distribution. We additionally propose two ways to generate pseudo anomalies: patch and skip frame based. Extensive experiments on three challenging video anomaly datasets demonstrate the effectiveness of our method in improving conventional AEs, achieving state-of-the-art performance. 
\end{abstract}
\section{Introduction}
\label{sec:introduction}

Anomalous event detection in video sequences has recently attracted significant attention \cite{sultani2018real,zaheer2020claws,liu2018future,chang2020clustering,li2013anomaly,luo2017revisit,lu2013abnormal,abati2019latent,zhong2019graph}. The task is extremely challenging because, in real-life situations such as in surveillance videos, anomalous events do not occur frequently. Moreover, there is no restriction on the types of anomaly events that may occur, making it cumbersome to collect sufficient anomaly examples.
Therefore, anomaly detection is often seen as one-class classification (OCC) problem in which only normal data is used to train a novelty detection model \cite{gong2019memorizing,liu2018future,chang2020clustering,sabokrou2018adversarially,zaheer2020old,liu2008isolation,lee2012anomaly}. 

\begin{figure}
\begin{center}
\includegraphics[width=\linewidth]{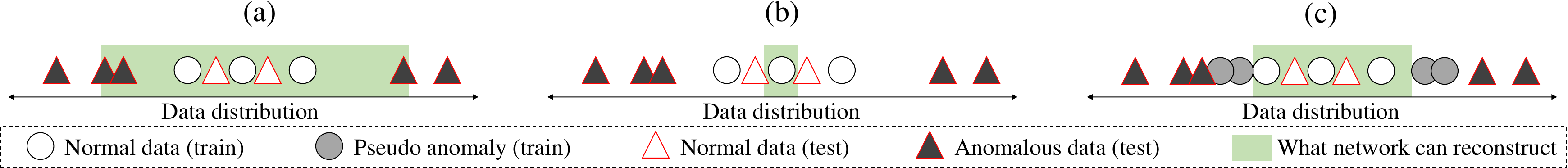}
\vspace{-10mm}
\end{center}
   \caption{An illustration of reconstruction capability of the three variants of AE: (a) conventional AE trained only on normal data, (b) methods that limit reconstruction of anomalous data without pseudo anomalies, and (c) our approach which encourages the network to reconstruct only normal data with the assistance of pseudo anomalies.}
\label{fig:motivationdatadistribution}
\end{figure}

One way to tackle the OCC problem is by using a deep autoencoder (AE) trained to reconstruct normal data 
\cite{hasan2016learning,zhao2017spatio,luo2017revisit,luo2017remembering,gong2019memorizing,park2020learning}. This way, the model is encouraged to encode normalcy information within its latent space. 
At test time, the trained AE is then expected to only reconstruct normal cases while failing to reconstruct the anomalous cases.
However, 
as reported in literature \cite{zong2018deep,munawar2017limiting,zaheer2020old,gong2019memorizing} as well as observed in our experiments (baseline performances in Fig. \ref{fig:qualitativeresults}), 
an AE can also often reconstruct anomalous examples. It is a likely outcome as the reconstruction boundary of the AE trained only on normal data would be unconstrained as long as the boundary includes the normal data in the training set.
Therefore, the reconstructions between normal and anomalous data may not be discriminative enough to successfully identify the anomalies. 
The phenomenon is illustrated in Fig. \ref{fig:motivationdatadistribution}(a).

To alleviate this problem, several researchers \cite{gong2019memorizing,park2020learning} proposed employing a memory mechanism over the latent space between the encoder and the decoder of an AE to limit the reconstruction capability in the case of anomalous input. The idea is to memorize normal representations learned from the training data. This way, the network is restricted to use the memorized normalcy representations for reconstruction, thus reducing its capability to regenerate anomalous data. 
However, such a network is highly dependent on the memory size and a small-sized memory may also limit its normal data reconstruction capability. For instance, Fig. 6 of \cite{gong2019memorizing} and Fig. \ref{fig:qualitativeresults} (MemAE) show that although anomalous regions have more distortions compared to the baseline, some of the normal portions are distorted as well, which may result in a limited discrimination between normal and anomalous data. The phenomenon can be attributed to the lack of a vivid reconstruction boundary when limiting the reconstruction using only normal data during training, thus resulting in a limited reconstruction capability for the normal data at test time, as illustrated in Fig. \ref{fig:motivationdatadistribution}(b). 

In this work, we also propose to limit an AE in reconstructing anomalies; however, in such a way that the normal reconstructions are not affected.
Particularly, we introduce a novel training mechanism of an AE with the objective of reconstructing only normal data even if the input is anomalous.
Since there are no real anomalies in the training data under the OCC setting, we propose the idea of generating and utilizing pseudo anomalies to assist the training. To this end, two types of pseudo anomaly generation methods are explored, i.e., patch and skip frame based, 
to simulate out of normal data distribution from normal data.
By encouraging to reconstruct only normal data for any kind of input (i.e., normal or pseudo anomalous), 
AEs are specifically trained to limit their reconstruction boundaries around the normal data hence not affecting the normal reconstructions while distorting anomalies, as illustrated in Fig. \ref{fig:motivationdatadistribution}(c). This results in an improved discrimination between normal and anomalous data, which is evident from the superior performance of our approach both qualitatively and quantitatively.

The contributions of this work are threefold: 1) We propose a pseudo anomaly based novel method of encouraging only normal data reconstructions to train AEs in the OCC setting. 2) We propose two types of pseudo anomalies, patch and skip frame based, to simulate anomalies. 3) In the experiments, we present extensive evaluations and analysis of the proposed training approach using each of the  pseudo anomaly types on three challenging video anomaly detection datasets including Ped2 \cite{li2013anomaly}, Avenue \cite{lu2013abnormal}, and ShanghaiTech \cite{luo2017revisit}.

\vspace{-4mm}
\section{Related Works}
\label{sec:relatedworks}
\vspace{-3mm}



\noindent{\textbf{Reconstruction Based Methods:}} A common way to tackle the one-class classification (OCC) problem is by utilizing autoencoders (AEs) which learn normal data representations by reconstructing the inputs \cite{hasan2016learning,zhao2017spatio,luo2017revisit,luo2017remembering,gong2019memorizing,park2020learning}.
However, since AEs can also well-reconstruct anomalous data \cite{zong2018deep,munawar2017limiting,zaheer2020old,gong2019memorizing}, several researchers proposed memory based networks to limit reconstruction capability of AEs \cite{gong2019memorizing,park2020learning}. The idea is to use only the learned memory vectors for reconstruction, which helps in achieving higher reconstruction loss for anomalous inputs.
However, such a configuration may also restrain the normal data reconstruction capability due to its limited memory size.
In contrast, our approach encourages AEs to produce unconstrained reconstructions for normal inputs while limiting the reconstructions for anomalous inputs, thus producing more discriminative anomaly scores.

\noindent{\textbf{Non-Reconstruction Methods:}} 
Several researchers adopt different schemes for OCC based anomaly detection: focusing only on objects by utilizing object detectors in the frameworks \cite{hinami2017joint,ionescu2019object,doshi2020any,doshi2020continual,sun2020scene,yu2020cloze,georgescu2021background}; predicting future frames from the past few consecutive frames with the intuition that it is difficult to predict unseen anomalous data \cite{liu2018future,park2020learning,lu2019future,lu2020few,dong2020dual};
or incorporating adversarial components \cite{ravanbakhsh2017abnormal,lee2019bman,liu2018future,vu2019robust,ji2020tam,lee2018stan}. 
Our approach is different as we do not utilize any additional component and solely rely on the reconstruction based AEs.

\noindent{\textbf{Pseudo Anomalies:}} There have been a few recent attempts towards pseudo anomaly generation for one-class classifiers. Georgescu \etal \cite{georgescu2021background} utilize time magnification and separate datasets as pseudo abnormal objects to train an object-centric architecture by flipping the gradient for pseudo abnormal objects.
However, this approach can be only applicable to anomalies related to objects and requires a pretrained object detectors.
OGNet \cite{zaheer2020old} and G2D \cite{pourreza2021g2d} propose using an under-trained and adversarially learned generator for generating fake anomaly data to train a binary classifier. Furthermore, OGNet, passes a fusion of two images to a fully-trained generator to produce another type of fake anomalous example.
These approaches require a two-phase training, one for adversarial training of generator and the other phase for training binary classifier.
Differently, our approach is not restricted to any predefined object classes, carries out the training in an end-to-end manner, and does not require any pretrained networks. 

\noindent{\textbf{Data Augmentation:}} Pseudo anomaly generation used in our method can also be viewed as a form of data augmentation technique, widely popular among image classification models, which manipulates training data to increase variety \cite{bengio2011deep,krizhevsky2012imagenet,yun2019cutmix,lee2020smoothmix,zhang2018mixup}. Typically, the class labels for the augmented data are derived from the already exiting classes in the dataset. In contrast, generating pseudo anomalies can be seen as augmenting data into a new class, i.e., anomaly, which is not a part of the already existing classes. 

\noindent{\textbf{Non-OCC Methods:}} In order to enhance the discrimination between normal and anomalous data, some researchers \cite{munawar2017limiting, yamanaka2019autoencoding} propose to deviate from the fundamental definition of OCC by using real anomaly examples during training. We also acknowledge a recent introduction of several weakly supervised methods using video-level binary annotations for training \cite{sultani2018real,zaheer2020claws,zhong2019graph}. 
However, our approach is not directly comparable to these approaches as we do not train using any real anomaly examples. Instead, our method can be categorized as OCC because we utilize only normal training data to synthesize pseudo anomaly examples.

\vspace{-4mm}
\section{Methodology}
\label{sec:methodology}
\vspace{-3mm}

The overall configuration of our approach (Fig. \ref{fig:overall}) consists of a reconstruction based autoencoder (AE) along with its training objective and a pseudo anomaly generator. Each component is discussed next:

\begin{figure}
\begin{center}
\includegraphics[width=\linewidth]{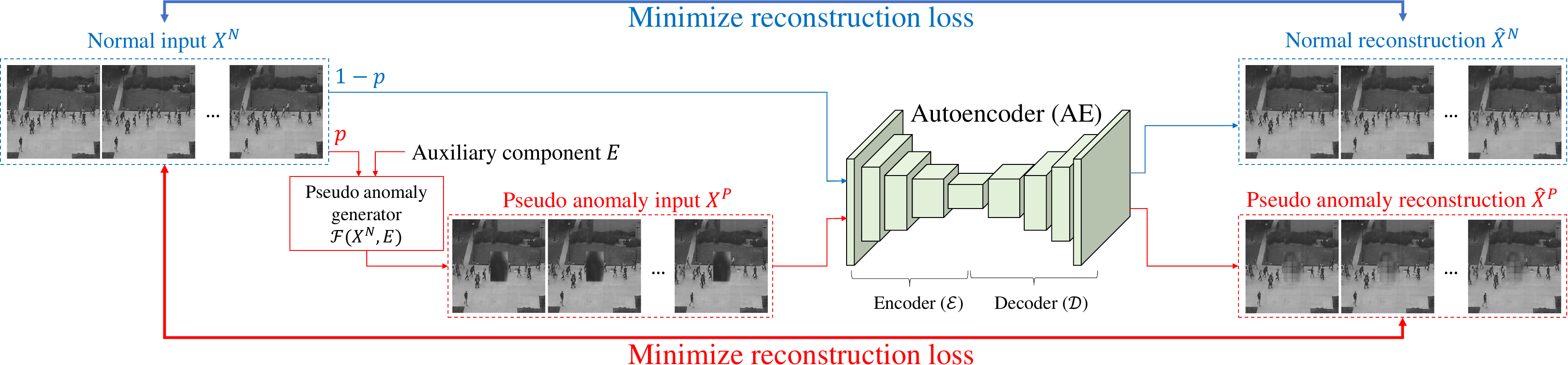}
\vspace{-8mm}
\end{center}
   \caption{We train an AE for OCC using normal as well as pseudo anomaly sequences. Pseudo anomalies are generated using normal data and extra components to simulate anomalies. With a probability $p$, pseudo anomalies are introduced in the training examples for the AE. 
   The AE is trained to not to reconstruct anomalies by encouraging to reconstruct the corresponding normal training examples from which the pseudo anomaly is generated.
   }
\label{fig:overall}
\vspace{-2mm}
\end{figure}

\vspace{-2mm}
\subsection{Conventional Autoencoders in OCC Setting}
\label{sec:trainingconventionalae}

For an input $X$, an AE can be defined as:
\begin{equation}
    \hat{X} = \mathcal{D}(\mathcal{E}(X)) \text{,}
\label{eq:ae}
\end{equation}
where $\mathcal{E}$ and $\mathcal{D}$ are encoder and decoder networks, respectively. 
The encoder generates a latent code of a typically smaller dimension compared to the input. This code is then transformed into the reconstruction $\hat{X}$ of the input by the decoder. In order to capture rich information from video data, AEs are often designed to take multiple frames as input \cite{gong2019memorizing,park2020learning,hasan2016learning,zhao2017spatio}. Following this convention, we also set our AE model to take $X$ as input of size $T \times C \times H \times W$, where $T$, $C$, $H$, and $W$ are the number of frames, number of channels, height, and width of the frames in the input sequence, respectively. 

Typically, to tackle OCC problem, AEs are utilized to learn the normal representations by
minimizing the reconstruction loss between the normal input $X^N$ and its reconstruction $\hat{X}^N$ as follows:
\begin{equation}
    L^N= \frac{1}{T \times C \times H \times W}  \left \| \hat{X}^N - X^N  \right \|_{F}^{2} \text{,}
\label{eq:aereconloss}
\end{equation}
where $\left \| .  \right \|_{F}$ is Frobenius norm.
With this training setting, an AE is ideally expected to reconstruct only normal data while unable to reconstruct anomalous data. However, as widely reported across the literature, AE can often ``generalize" too well and start reconstructing anomalous examples as well \cite{zong2018deep,munawar2017limiting,zaheer2020old,gong2019memorizing}. We try to encourage AE to produce only normal data reconstructions for both normal and anomalous inputs,
hence the network is unable to reconstruct anomalies at test time.

\subsection{Learning Not to Reconstruct Anomalies}

We propose a training mechanism with the objective to encourage an AE towards reconstructing only normal data regardless of the input. This means even if the data contain an abnormality, the network will learn to produce a normal reconstruction. In OCC setting, as we do not have access to real anomalous examples during training, we utilize pseudo anomalies $X^P$, which are 
generated by altering normal data $X^N$ as: 
\begin{equation}
    X^P = \mathcal{F}(X^N, E) \text{,}
\label{eq:generatingpseudoanomaly}
\end{equation}
where $\mathcal{F}(\cdot)$ is a pseudo anomaly generator function and $E$ is an auxiliary component defined based on the intended pseudo anomaly type. Both $\mathcal{F}(\cdot)$ and $E$ are discussed in Section \ref{subsec:generatingpseudoanomalies}.


To train the network, we utilize $X^P$ as input with a probability $p$ and $X^N$ with a probability $1-p$, where the hyperparameter $p$ controls the ratio of pseudo anomalies.
Given normal and pseudo anomalous data as part of the training,
the loss of the network is then defined as:
\begin{equation}
    L= 
\begin{cases}
    L^{N} & \text{if } X=X^{N} \text{,}\\
    L^{P} & \text{if } X=X^{P} \text{,}\\
\end{cases}
\label{eq:reconloss}
\end{equation}
where $L^N$ is the reconstruction loss for normal data as defined in Eq. \eqref{eq:aereconloss} and $L^{P}$ is the reconstruction loss for pseudo anomaly data given as:
\begin{equation}
    L^{P}=  \frac{1}{T \times C \times H \times W} \left \| \hat{X}^P - X^N  \right \|_{F}^{2}  \text{.}
\label{eq:pseudoreconloss}
\end{equation}
It may be noted that, both $L^N$ and $L^P$ encourage the network to reconstruct normal data. 
Specifically, by minimizing $L^P$ with respect to $X^N$, the network attempts to remove the perturbations introduced by pseudo anomaly generator.

\subsection{Generating Pseudo Anomalies}
\label{subsec:generatingpseudoanomalies}
In this section, we discuss two distinct methods to generate pseudo anomalies from normal data. 
Specifically, we formally define the pseudo anomaly generator function $\mathcal{F}(\cdot)$ and its auxiliary component $E$ (Eq. \eqref{eq:generatingpseudoanomaly}) for each pseudo anomaly category. 


\vspace{-2mm}
\subsubsection{Patch Based Pseudo Anomalies}
\label{subsubsec:patchbasedpseudoanomalies}

In real world scenarios, an OCC based anomaly detection system may encounter cases like unusual objects. Derived from this motivation, we propose 
to overlay all normal input frames $X^N$ with an anomalous patch $A$ (i.e., $E=A$), which
is essentially taken from an arbitrary image $I^A$ from some other dataset, e.g., CIFAR-100 \cite{krizhevsky2009learning}, referred as intruding dataset in this paper. Since such overlaid frames are not actually a part of the normal data, these are anomalous. By default, each patch is placed using SmoothMixS \cite{lee2020smoothmix}, which smooths out the boundary, to prevent the network from latching on to the edges of the patch.
Nonetheless, the performance of our method is not strictly dependent on any particular patching technique, as discussed in Section \ref{subsubsec:hyperparameterseval}.

To generate the $i$-th frame of pseudo anomaly $X^{P}_i$ in the input sequence, we first transform $I^A$ taken from the intruding dataset to $C \times H \times W$, same as the input frame size.
Then, using SmoothMixS mask, a patch $A$ of size $(\sigma^w, \sigma^h)$ is extracted from $I^A$ with $(\mu_i^x,\mu_i^y)$ defining its center coordinates.
$A$ is then
overlaid on the $i$-th frame $X^N_i$ in the normal sequence.
The center coordinates $(\mu_i^x,\mu_i^y)$ are randomly selected within the image dimensions whereas the patch size $(\sigma^w, \sigma^h)$ is randomly selected from 10 pixels to $\alpha W$ for the patch width ($\sigma^w \in [10, \alpha W]$) and from 10 pixels to $\alpha H$ for the patch height ($\sigma^h \in [10, \alpha H]$).  $\alpha$ is a hyperparameter to adjust the maximum size of the patch.
Detailed visualization of the process can be seen in Fig. 1 of the Supplementary. Furthermore, this technique can be generalized to utilizing a video dataset as an intruder dataset by using $T$ frames, i.e., $I^A = (I^A_1, I^A_2, ..., I^A_T)$, with each time step corresponding to the respective time step of the input.

\begin{figure}
\begin{center}
\includegraphics[width=\linewidth]{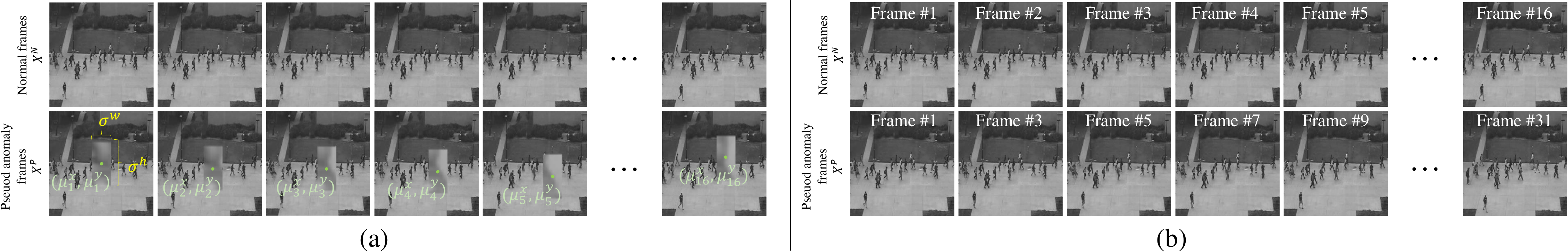}
\vspace{-10mm}
\end{center}
   \caption{Examples of normal data and the corresponding pseudo anomalies generated using (a) patch and (b) skip frame based methods ($s=2$).}
\label{fig:pseudoanomaly}
\vspace{-4mm}
\end{figure}

Moreover, to incorporate movement in pseudo anomalies, the position of a given patch in a particular frame within the sequence is changed based on the previous position as:
\begin{equation}
    \mu_{i}^x = \mu_{i-1}^x + \Delta\mu_i^x \text{,} \;\;\;\;\;\;
    \mu_{i}^y = \mu_{i-1}^y + \Delta\mu_i^y \text{, }
\label{eq:smoothmix_mask}
\end{equation}
where $\Delta\mu_i^x$ and $\Delta\mu_i^y$ are each randomly selected within the range of $-\beta$ to $\beta$ for every $i$ and $\beta$ is the hyperparameter to adjust the maximum movement of the patch in terms of pixels. An example of patch based pseudo anomaly along with its corresponding normal data can be seen in Fig. \ref{fig:pseudoanomaly}(a).


\vspace{-2mm}
\subsubsection{Skip Frame Based Pseudo Anomalies}
\label{subsubsec:skipframebasedpseudoanomalies}

In addition to the anomalous objects as described in Section \ref{subsubsec:patchbasedpseudoanomalies}, in real-world scenarios, anomalies may sometimes look normal in their appearance while depicting anomalous movements.
To create a system that is inclusive of this scenario, similar in essence with motion magnification of \cite{georgescu2021background}, we propose skipping frames to generate anomalous movements from the normal training sequences. Given a video consisting of $K$ images $(I_1, I_2, ..., I_K)$, a normal temporally-consistent input frame sequence $X^N$ of length $T$ is taken from the video starting from a random $n$-th index as follows:
\begin{equation}
    X^N = (I_n, I_{n+1}, ..., I_{n+(T-1)}) = (I_{n+t})_{0 \leq t < T} \text{. }
\label{eq:skipframenormal}
\end{equation}
To generate the corresponding pseudo anomaly $X^P$, we take the first frame of $X^N$, then replace its following frames with skipped frames $E=(I_{n+s}, I_{n+2s}, ..., I_{n+(T-1)s})$ as: 
\begin{equation}
\begin{split}
    X^P = (I_n, E) = (I_n, I_{n+s}, ..., I_{n+(T-1)s}) = (I_{n+ts})_{0 \leq t < T, s>1} \text{, }
\label{eq:sequencetemporalpseudo}
\end{split}
\end{equation}
where $s$ is a hyperparameter controlling the number of skipped frames. 
An example of a pseudo anomaly generated with skip frame method and its corresponding normal sequence can be seen in Fig. \ref{fig:pseudoanomaly}(b).

\vspace{-2mm}
\subsection{Inference}
At test time, we process the input sequences and their anomaly scores. Concurrent with other recent anomaly detection methods \cite{park2020learning,liu2018future,dong2020dual}, we utilize Peak Signal to Noise Ratio (PSNR) $P_t$ between an input frame and its reconstruction to compute the anomaly score as follows:
\begin{equation}
    P_t = 10 \text{ log}_{10}  \frac{M_{\hat{I}_t}^2}{\frac{1}{R} \left \| \hat{I}_t - I_t  \right \|_{F}^{2} } \text{,}
\label{eq:psnr}
\end{equation}
where $t$ is the frame index, $I_t$ is the $t$-th frame input, $\hat{I}_t$ is the reconstruction of $I_t$, $R$ is the total number of pixels in $\hat{I}_t$, and $M_{\hat{I}_t}$ is the maximum possible pixel value of $\hat{I}_t$. 
Finally, following \cite{park2020learning,liu2018future,dong2020dual}, the anomaly score $S_t$ is obtained using min-max normalization of $P_t$ as:
\begin{equation}
    S_t = 1 - \frac{P_t - \min_t(P_t)}{\max_t(P_t)-\min_t(P_t)} \text{,}
\label{eq:anomalyscore}
\end{equation}
where a higher $S_t$ value represents higher reconstruction error compared to the other frames in the test video and vice versa.

\vspace{-3mm}
\section{Experiments}
\label{sec:experiments}
\vspace{-2mm}


\vspace{-1mm}
\subsection{Experimental Setup}
\label{subsec:experimentalsetup}
\vspace{-1mm}
\noindent\textbf{Datasets.} We evaluate our approach on three publicly available video anomaly detection datasets, i.e., Ped2 \cite{li2013anomaly}, Avenue \cite{lu2013abnormal}, and ShanghaiTech \cite{luo2017revisit}. We utilize the standard division of the datasets in which training splits consist of only normal videos. Whereas, every video in each of the test sets contains one or more anomalous portions. Further details about the datasets are provided in the Supplementary.

\noindent\textbf{Evaluation Criteria.}
We evaluate our approach using the widely popular frame-level area under the ROC curve (AUC) metric \cite{zaheer2020old}. The ROC curve is obtained by varying the anomaly score thresholds to plot false and true positive rates across the whole test set, i.e., one ROC curve for a dataset. Higher AUC values represent more accurate results.

\noindent\textbf{Parameters and Implementation Details.}
For the AE architecture, we use a 3D convolution-deconvolution network similarly proposed by Gong \etal \cite{gong2019memorizing}. The AE takes an input sequence $X$ (Eq. \eqref{eq:ae}) of size $16 \times 1 \times 256 \times 256$ and produces its reconstruction of the same size. During training, the reconstruction loss is calculated across all of the 16 frames. At test time, only the 9th frame of a sequence is considered for anomaly score calculation (Eq. \eqref{eq:psnr} - \eqref{eq:anomalyscore}). Further details on the architecture are provided in the Supplementary. 
By default, for all datasets, we utilize $p=0.2$, $s=\{2,3,4,5\}$, $\alpha=0.5$, CIFAR-100 \cite{krizhevsky2009learning} as the intruder dataset, and SmoothMixS \cite{lee2020smoothmix} as the patching technique. Moreover, $\beta$ is set to $10$ for Ped2 and $25$ for the other datasets. $s=\{2,3,4,5\}$ means $s$ is randomly selected as $2$, $3$, $4$ or $5$ each time we generate pseudo anomaly. 
To observe the robustness of our method, in Section \ref{subsubsec:hyperparameterseval}, we also perform evaluations by varying hyperparameters.
Training is carried out separately for the model trained without pseudo anomalies (referred as baseline; see Section \ref{sec:trainingconventionalae}), the model trained using patch based pseudo anomalies, and the model trained using skip frame based pseudo anomalies. 
The code is provided at \textit{https://github.com/aseuteurideu/LearningNotToReconstructAnomalies}.


\vspace{-3mm}
\subsection{Quantitative Results}
\label{subsec:sota}

\vspace{-1mm}
\subsubsection{Comparisons with the Baseline and SOTA Methods}
\vspace{-1mm}
Table \ref{tab:sota} shows the AUC comparisons of our overall model with the existing state-of-the-art (SOTA) approaches on Ped2 \cite{li2013anomaly}, Avenue \cite{lu2013abnormal}, and ShanghaiTech \cite{luo2017revisit} datasets. 
For a fair comparison, we classify various SOTA methods into five categories: 1) Non-deep learning approaches, 2) Object-centric methods which utilize object detectors to focus only on the detected objects, 3) Prediction based methods that predict a future frame to detect anomalies, 4) Reconstruction based approaches that use reconstruction of the input to detect anomalies, and 5) Miscellaneous methods which are either the tasks not belonging to the aforementioned categories or employing a combination of these. Our method falls in the category of reconstruction based methods. 

Comparing to the other approaches of the same category, i.e., reconstruction, our model achieves the best performance on all three benchmark datasets. Interestingly, both of our models trained with different kinds of pseudo anomalies achieve better performance than the memory based networks, such as MNAD-Recon \cite{park2020learning} and MemAE \cite{gong2019memorizing}, considering that we use a very similar network architecture with these approaches and a common goal of limiting the AE capability of reconstructing anomalies. It may also be noted that our proposed pseudo anomaly based trained models provide consistent gains over the respective baselines on all three datasets. 
This clearly demonstrates the superiority of our proposed approach, i.e., training AEs by encouraging only normal data reconstructions assisted by pseudo anomalies. 

Looking at the techniques in other categories, our proposed approach demonstrates a comparable performance.
Particularly, compared with the architectures that are designed with complex components, such as attention, optical flow, adversarial training, LSTM, etc., for example, in BMAN \cite{lee2019bman}, our method provides an overall comparable performance without any bells and whistles. In addition, most methods in the object-centric category \cite{ionescu2019object,doshi2020any,doshi2020continual,sun2020scene,yu2020cloze,georgescu2021background} require pre-trained object detectors which make their applicability limited to the set of predefined object categories. 
In contrast, while our method is generic and can be applied to a variety of AE based architectures, it is also not constrained by object detectors.


\begin{table}[]

\resizebox{\linewidth}{!}{
\centering
\begin{tabular}[t]{c|l|ccc|}
\hline
\multicolumn{2}{c|}{Methods}                  & Ped2 \cite{li2013anomaly}  & Ave \cite{lu2013abnormal} & Sh \cite{luo2017revisit}     \\ \hline \hline
\parbox[t]{2mm}{\multirow{18}{*}{\rotatebox[origin=c]{90}{Miscellaneous}}}           
  & AbnormalGAN \cite{ravanbakhsh2017abnormal}     & 93.5\%  & -       & -       \\
  & Smeureanu \etal \cite{smeureanu2017deep}       & -       & 84.6\%  & -       \\
  & AMDN \cite{xu2015learning,xu2017detecting}                    & 90.8\%  & -       & -       \\
  & STAN \cite{lee2018stan}                        & 96.5\%  & 87.2\%  & -       \\
  & MC2ST \cite{liu2018classifier}                 & 87.5\%  & 84.4\%  & -       \\
  & Ionescu \etal   \cite{ionescu2019detecting}    & -       & 88.9\%  & -       \\
  & BMAN \cite{lee2019bman}                        & 96.6\%  & \textbf{90.0\%}  & \underline{76.2\%}  \\
  & AMC  \cite{nguyen2019anomaly}                  & 96.2\%  & 86.9\%  & -       \\
  & Vu \etal \cite{vu2019robust}                   & \textbf{99.21\%}  & 71.54\%  & -       \\
  & DeepOC   \cite{wu2019deep}                     & -     & 86.6\%  & -       \\
  & TAM-Net  \cite{ji2020tam}                      & \underline{98.1\%}  & 78.3\%  & -       \\
  & LSA \cite{abati2019latent}                     & 95.4\%  & -       & 72.5\%  \\
  & Ramachandra \etal \cite{ramachandra2020learning} & 94.0\%  & \underline{87.2\%}  & -   \\
  & Tang \etal \cite{tang2020integrating}          & 96.3\%  & 85.1\%  & 73.0\%  \\
  & Wang \etal \cite{wang2020cluster}              & -       & 87.0\%  & \textbf{79.3\%}  \\
  & OGNet \cite{zaheer2020old}                     & \underline{98.1\%}  & -       & -       \\
  & Conv-VRNN \cite{lu2019future}         & 96.06\%  & 85.78\%  & -  \\ 
  & Chang \etal \cite{chang2020clustering}         & 96.5\%  & 86.0\%  & 73.3\%  \\ \hline
\parbox[t]{2mm}{\multirow{6}{*}{\rotatebox[origin=c]{90}{Object-centric}}}   
  & MT-FRCN \cite{hinami2017joint}                     & 92.2\%  & -       & -       \\
  & Ionescu \etal \cite{ionescu2019object} \footnotemark             & 94.3\%  & 87.4\%  & \underline{78.7\%}  \\
  & Doshi and Yilmaz \cite{doshi2020any,doshi2020continual} & \underline{97.8\%}  & 86.4\%  & 71.62\%  \\
  & Sun \etal \cite{sun2020scene}                      & -       & \underline{89.6\% } & 74.7\%  \\
  & VEC \cite{yu2020cloze}                             & 97.3\%  & \underline{89.6\% } & 74.8\%  \\ 
  & Georgescu \etal \cite{georgescu2021background}             & \textbf{98.7\%}  & \textbf{92.3\%}  & \textbf{82.7\%} \\ \hline

\end{tabular}
\begin{tabular}[t]{c|l|ccc|}
\hline
\multicolumn{2}{c|}{Methods}                  & Ped2 \cite{li2013anomaly}  & Ave \cite{lu2013abnormal} & Sh \cite{luo2017revisit}     \\ \hline \hline
\parbox[t]{2mm}{\multirow{10}{*}{\rotatebox[origin=c]{90}{Non deep learning}}}
  & MPPCA \cite{kim2009observe}                        & 69.3\%  & -       & -     \\
  & MPPC+SFA \cite{kim2009observe}                     & 61.3\%  & -       & -     \\
  & Mehran \etal \cite{mehran2009abnormal}             & 55.6\%  & -       & -     \\
  & MDT   \cite{mahadevan2010anomaly}                  & 82.9\%  & -       & -     \\
  & Lu \etal \cite{lu2013abnormal}                     & -       & \textbf{80.9\%}  & -     \\
  & AMDN   \cite{xu2017detecting}                      & \underline{90.8\%}  & -       & -     \\
  & Del Giorno \etal \cite{del2016discriminative}      & -       & \underline{78.3\%}  & -     \\
  & LSHF    \cite{zhang2016video}                      & \textbf{91.0\%}  & -       & -     \\
  & Xu \etal \cite{xu2014video} & 88.2\%  & -  & -     \\
  & Ramachandra and Jones \cite{ramachandra2020street} & 88.3\%  & 72.0\%  & -     \\ \hline
\parbox[t]{2mm}{\multirow{4}{*}{\rotatebox[origin=c]{90}{Prediction}}}        
  & Frame-Pred  \cite{liu2018future}          & 95.4\%  & 85.1\%  & 72.8\%  \\
  & Dong \etal \cite{dong2020dual}            & 95.6\%  & 84.9\%  & \underline{73.7\%}  \\
  & Lu \etal \cite{lu2020few}                 & \underline{96.2\%}  & \underline{85.8\%}  & \textbf{77.9\%}  \\
  & MNAD-Pred \cite{park2020learning}         & \textbf{97.0\%}  & \textbf{88.5\%}  & 70.5\%  \\ \hline
\parbox[t]{2mm}{\multirow{10}{*}{\rotatebox[origin=c]{90}{Reconstruction}}}   
  & AE-Conv2D  \cite{hasan2016learning}          & 90.0\%  & 70.2\%  & 60.85\% \\
  & AE-Conv3D  \cite{zhao2017spatio}             & 91.2\%  & 71.1\%  & -       \\
  & AE-ConvLSTM  \cite{luo2017remembering}       & 88.10\% & 77.00\% & -       \\
  & TSC \cite{luo2017revisit}                    & 91.03\% & 80.56\% & 67.94\% \\
  & StackRNN \cite{luo2017revisit}               & 92.21\% & 81.71\% & 68.00\% \\
  & MemAE \cite{gong2019memorizing}              & 94.1\%  & 83.3\%  & 71.2\%  \\
  & MNAD-Recon \cite{park2020learning}           & 90.2\%  & 82.8\%  & 69.8\%  \\ \cdashline{2-5} 
  & Baseline              & 92.49\%  & 81.47\% & 71.28\% \\
  & \textbf{Ours}: Patch based      &  \underline{94.77\%} & \textbf{84.91\%} & \underline{72.46\%} \\
  & \textbf{Ours}: Skip frame based      & \textbf{96.50\%} & \underline{84.67\%} & \textbf{75.97\%} \\ \hline
\end{tabular}
}
\vspace{-2mm}
\footnotemark[\value{footnote}]\footnotesize{Micro AUC reported in \cite{georgescu2021background}}
\caption{AUC performance comparison of our approach with several existing SOTA methods on Ped2, Avenue (Ave), and ShanghaiTech (Sh). Best and second best performances are highlighted as bold and underlined, in each category and dataset.
}
\label{tab:sota}
\vspace{-3mm}
\end{table}

\begin{figure}
\begin{center}
\includegraphics[width=\linewidth]{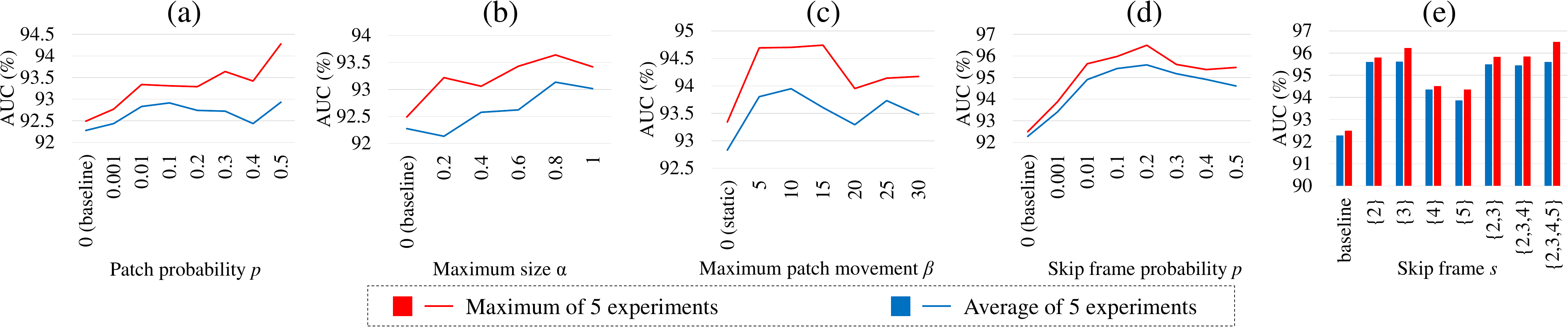}
\end{center}
\vspace{-4mm}
   \caption{
   Evaluations to show the robustness of our method on a wide-range of hyperparameters values:
   (a) patch probability $p$ given $\alpha=0.5$ and $\beta=0$; (b) maximum patch size $\alpha$ given $p=0.01$ and $\beta=0$; (c) maximum patch movement $\beta$ given $p=0.01$ and $\alpha=0.5$; (d) skip frame probability $p$ given $s=\{2,3,4,5\}$; (e) skip frame parameter $s$ given $p=0.2$.}
\label{fig:hyperparameterevaluation}
\vspace{-4mm}
\end{figure}

\begin{figure}
\begin{center}
\includegraphics[width=\linewidth]{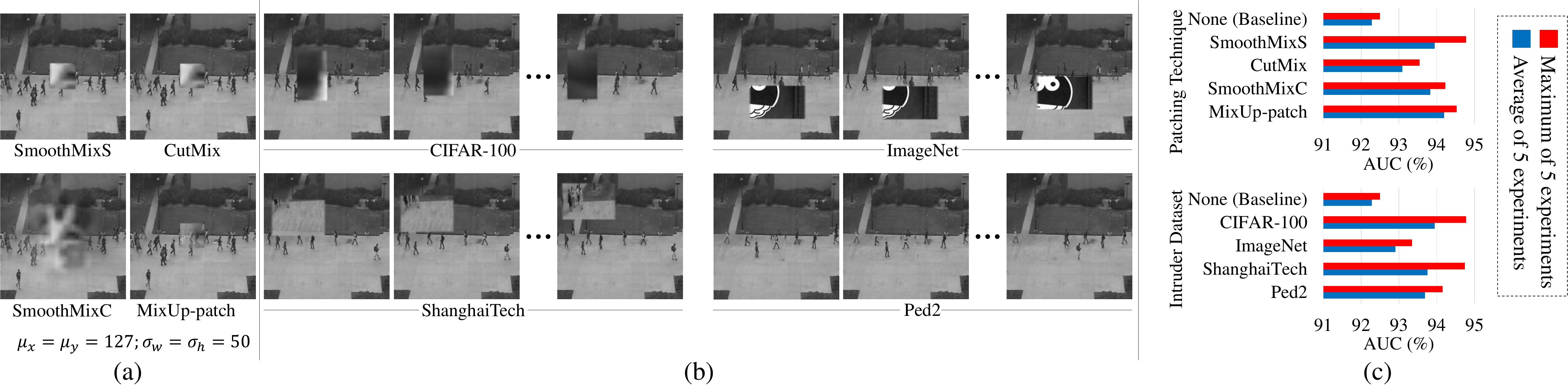}
\end{center}
\vspace{-5mm}
   \caption{(a) Pseudo anomalous frame samples generated using different patch techniques: SmoothMixS \cite{lee2020smoothmix}, CutMix \cite{yun2019cutmix}, SmoothMixC \cite{lee2020smoothmix}, and MixUp-patch (MixUp \cite{zhang2018mixup} with CutMix patch); (b) Visualizations of patch based pseudo anomaly sequence with SmoothMixS and different intruder datasets; (c) AUC comparisons between different patch techniques and intruder datasets.}
\label{fig:patchandintruderevaluation}
\vspace{-4mm}
\end{figure}

\vspace{-2mm}
\subsubsection{Hyperparameters Evaluation}
\label{subsubsec:hyperparameterseval}
To evaluate the robustness against the hyperparameters introduced in this work, we provide extensive analysis in Fig. \ref{fig:hyperparameterevaluation} and Fig. \ref{fig:patchandintruderevaluation}. Only Ped2 is used to limit the span of experiments. 

Fig. \ref{fig:hyperparameterevaluation}(a)-(c) show the evaluations for the hyperparameters used in patch based pseudo anomalies. 
Typically, the model trained using static patch based pseudo anomalies ($\beta=0$) successfully outperforms the baseline (Fig. \ref{fig:hyperparameterevaluation}(a)-(b)) and moving the patch location  ($\beta>0$) further improves the overall performance (Fig. \ref{fig:hyperparameterevaluation}(c)).
Additionally, we experiment using different patching techniques and intruder datasets. Pseudo anomalies generated using each of these methods can be seen in Fig. \ref{fig:patchandintruderevaluation}(a) \& (b). Fig. \ref{fig:patchandintruderevaluation}(c) shows the robustness of our method across different types of patching techniques and intruder datasets in outperforming the baseline. Interestingly, using Ped2 itself as an intruder dataset can also elevate the performance. It can be attributed to the anomalous shapes remained after cropping normal objects as we take patches from the Ped2 (Fig. \ref{fig:patchandintruderevaluation}(b)). Note that, we utilize only training sets of the intruder datasets. More details on patching techniques are provided in the Supplementary. 

Evaluations on a wide-range of hyperparameters used in skip frame based pseudo anomalies, i.e., $p$ and $s$, can be seen in Fig. \ref{fig:hyperparameterevaluation}(d)-(e). All experiments show performance gains over the baseline, with maximum performance achieved with $p=0.2$ and $s=\{2,3,4,5\}$.

\vspace{-3mm}
\subsection{Qualitative Results}
\label{subsec:qualitative}

\begin{figure}
\begin{center}
\includegraphics[width=\linewidth]{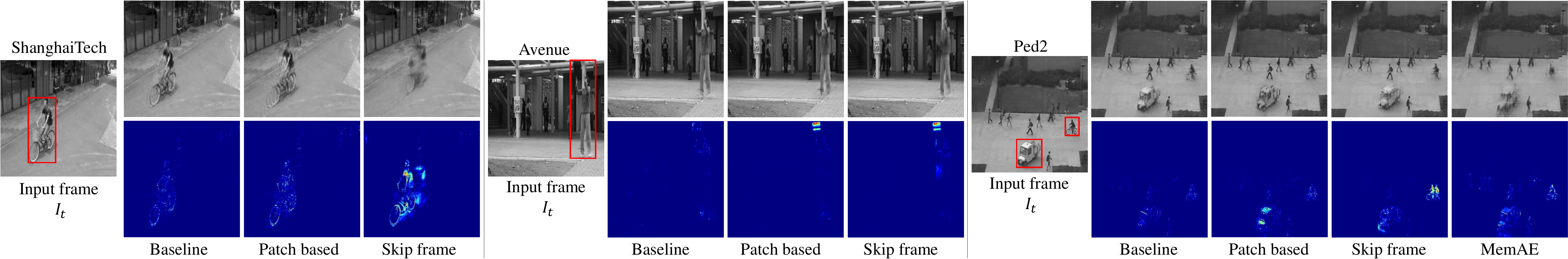}
\vspace{-10mm}
\end{center}
   \caption{
   Visualizations of input test frames, reconstructions (first row), and reconstruction error heatmaps (second row) of the baseline, our model trained using patch based pseudo anomalies, and our model trained using skip frame based pseudo anomalies on ShanghaiTech, Avenue, and  Ped2. Additionally, we provide comparison with MemAE \cite{gong2019memorizing} on Ped2. The anomalous regions are marked with red boxes. 
   }
\label{fig:qualitativeresults}
\vspace{-1mm}
\end{figure}

For a deeper understanding on how our method improves the baseline, in this section, we provide a qualitative comparisons of the baseline, our models, and MemAE \cite{gong2019memorizing}.
Fig. \ref{fig:qualitativeresults} shows several input test images from the three datasets, the reconstructions produced by different variants of our model, and the reconstruction error heatmaps. 
The heatmaps are generated by computing the squared error of each pixel between the input frame and its reconstruction, followed by min-max normalization. Based on the reconstruction error visualizations, it can be observed that both of our models successfully highlight the anomalous region more than the baseline by well reconstructing the normal data and poorly reconstructing the anomalous data, which results in the superior anomaly discrimination capability of our models. 
Furthermore, a few interesting observations may as well be noted. 
In the patch based method examples, the backpack being tossed up in Avenue is completely vanished from the reconstructions. Instead, our model reconstructs normal background. Similarly, reconstructions of bicycle riders produced by skip frame based model in ShanghaiTech and Ped2 are dislocated as if they were moving at a normal pedestrian pace. This demonstrates that, concurrent to our training objective, the learned model attempts to reconstruct normal even if the input is anomalous.
More qualitative results are provided in the Supplementary.

Furthermore, compared with MemAE \cite{gong2019memorizing} on Ped2, our model retains the quality of normal reconstructions. As discussed in Section \ref{sec:introduction} \& \ref{sec:relatedworks}, in addition to limiting reconstructions of anomalous regions, attributed to limited memory size, MemAE may also limit the reconstructions of normal parts. Therefore, it is less discriminative. It may be noted that, as MemAE official code provides only Ped2 trained model, we compare using only this dataset.

\begin{figure}
\begin{center}
\includegraphics[width=\linewidth]{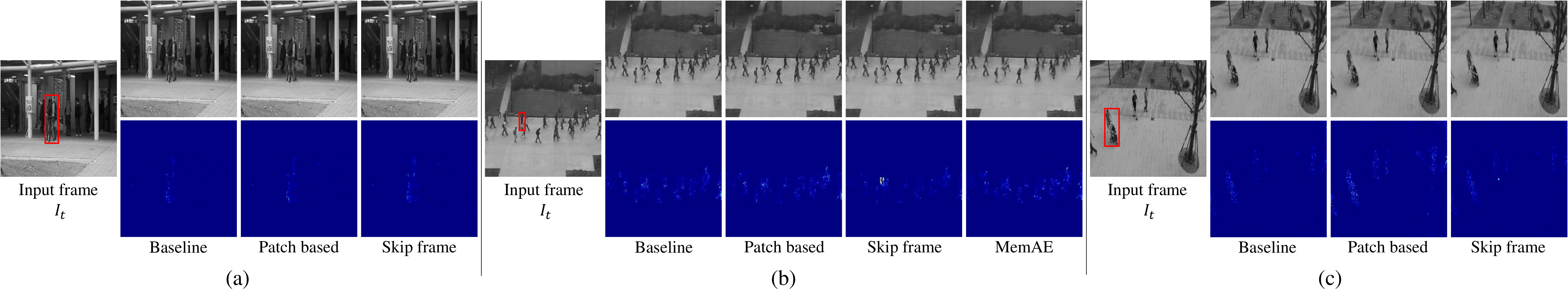}
\end{center}
\vspace{-7mm}
  \caption{
Input test frames, reconstructions (first row), and reconstruction error heatmaps (second row) on several examples when (a) both patch based and skip frame models failed, (b) patch based model failed, and (c) skip frame based model failed.
  }
\label{fig:failurecase}
\vspace{-3mm}
\end{figure}

To further understand the behavior of our models, Fig. \ref{fig:failurecase} shows various cases of failures produced by our model. Since the bicycle in Fig. \ref{fig:failurecase}(a) is too thin, both of our models, as well as the baseline, have difficulties in detecting the anomaly. Riding a skateboard in Fig. \ref{fig:failurecase}(b) is abnormal mainly because of its movement, but not because of its appearance as the skateboard is almost invisible. Our patch based pseudo anomaly model tends to have difficulties in such cases. The baseline and MemAE models also exhibit the same problem in this frame. Walking with a stroller in Fig. \ref{fig:failurecase}(c) is abnormal in its appearance, while the movement is normal. The skip frame based model tends to have the drawback in such case. Overall, these observations indicate that more careful choices of pseudo anomalies may lead to even better performances of the anomaly detection models.

\vspace{-3mm}
\section{Conclusion}
\label{sec:conclusion}

We propose a training mechanism of an autoencoder (AE) assisted by pseudo anomalies for one-class classification with the objective to reconstruct only normal data even if the input is not normal. This consequently increases the reconstruction error of anomalous inputs without restraining normal reconstructions, which leads to highly discriminative anomaly scores. 
To carry out this training, we additionally propose two pseudo anomaly generation methods, i.e., patch and skip frame based. Extensive evaluations on three challenging video anomaly datasets demonstrate that our proposed training methodology is effective for improving the capability of an AE to detect anomalies.  

\medskip
\noindent{\textbf{Acknowledgements}}. 
This  work  was supported by the Institute of Information \&  communications  Technology  Planning \&  Evaluation(IITP) grant  funded by the  Korea government (MSIT) (No.  2019-0-01309,  Development of AI  Technology for  Guidance of  a Mobile Robot to its  Goal  with  Uncertain Maps in Indoor/Outdoor Environments)

\bibliography{egbib}
\end{document}